\let\oldvec\vec
\documentclass[runningheads,a4paper]{llncs}
\let\vec\oldvec

\usepackage{algorithmic}
\usepackage{amssymb}
\usepackage{amsmath}
\usepackage{todonotes}
\usepackage{float}
\usepackage{subfig}

\usepackage{sistyle}
\SIthousandsep{,}

\usepackage{graphicx}

\usepackage{url}

\urldef{\mailsa}\path|vivek.mahato@ucdconnect.ie|
\urldef{\mailsb}\path|conrad.lee@ucd.ie|
\urldef{\mailsc}\path|padraig.cunningham@ucd.ie|    
\urldef{\mailscombined}\path|{firstname.lastname}@ucd.ie|

\newcommand{\keywords}[1]{\par\addvspace\baselineskip
\noindent\keywordname\enspace\ignorespaces#1}

\begin{document}
\hyphenation{name-space}

\mainmatter  
\title{A Case-Study on the Impact of Dynamic Time Warping in Time Series Regression}
\titlerunning{Dynamic Time Warping in Time Series Regression}
\author{Vivek Mahato\and P\'{a}draig Cunningham}
	\institute{School of Computer Science\\
	University College Dublin\\
	Dublin 4, Ireland\\
\mailsa, \mailsc}
%
%

%
%

\maketitle

\begin{abstract}It is well understood that Dynamic Time Warping (DTW) is effective in revealing similarities between time series that do not align perfectly. In this paper, we illustrate this on spectroscopy time-series data. We show that DTW is effective in improving accuracy on a regression task when only a single wavelength is considered. When combined with $k$-Nearest Neighbour, DTW has the added advantage that it can reveal similarities and differences between samples at the level of the time-series. 
However, in the problem, we consider here data is available across a spectrum of wavelengths. If aggregate statistics (means, variances) are used across many wavelengths the benefits of DTW are no longer apparent. We present this as another example of a situation where big data trumps sophisticated models in Machine Learning.

\end{abstract}

\section{Introduction}
In this paper, we consider the task of predicting the etch rate in silicon wafer production. While we frame this as a prediction task, our interest is in gaining an insight into the sources of process variability in plasma etching rather than very accurate prediction. Because we have insight as an objective we are interested in $k$-Nearest Neighbour ($k$-NN) classifiers because of their interpretability. 

In the production of silicon wafers, the etch rate is critically important because it directly influences the precision of the silicon structures. If a transistor gate length is too large the device speed is reduced, it if is too small it will leak. So it is important to be able to control the etch rate very precisely\cite{macgearailt2015process,flamm1981design}.

It is common to describe plasma etching as a cooking process or a recipe. As the recipe progresses different gases are introduced in the chamber to promote specific effects. Optical emission spectroscopy (OES) is the most widely used technique to track what is going on within the chamber \cite{shul1997high}. 

In the data we analyze here, the outcome is the etch rate, and the predictive variables are time-series spectra across 2048 wavelengths. The process takes about 42 seconds, the spectra are sampled at 1.3Hz so each of the 2048 OES time-series contains about 57 ticks \cite{macgearailt2015process}. With our objective of identifying sources of process variability, we are interested in identifying systematic differences in OES time-series that are predictive of differences in etch rate. 

In the next section, we introduce the problem domain and describe the data on which the analysis is based. In section \ref{sec:models} we describe the prediction models we use and we discuss the results of the evaluation in section \ref{sec:eval}. Some conclusions and directions for future work are presented in section \ref{sec:conc}.

\section{The Problem Domain}
Plasma etching is an integral part of the pipeline in wafer manufacturing. This process is conducted within specialised etch chambers where etchant gases in plasma form are directed towards the wafer surface where the exposed material is removed. This removal of material contributes to the formation of semiconductor components on the wafer \cite{flamm1981design,lynn2012global,lynn2012real}. 

The etching process involves chemical reactions that emit a spectrum of frequencies of electromagnetic radiation, which can be captured by a spectroscope. The captured spectroscopic data for each wafer is a high dimensional multivariate time series where each dimension represents one wavelength through a period of time during the etching process \cite{macgearailt2015process}.
\begin{figure}[H]
\includegraphics[width=330pt]{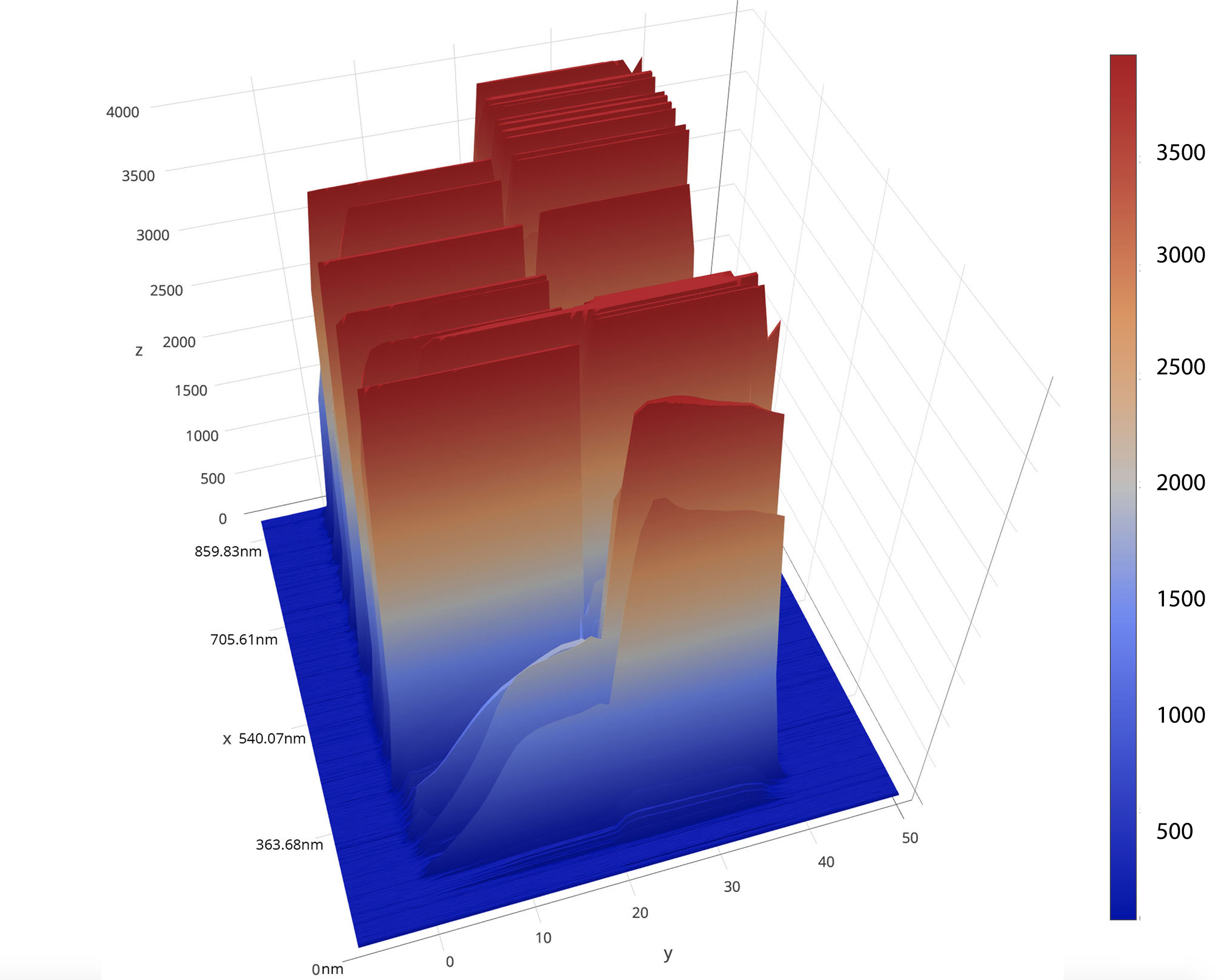}
\caption{Spectroscopic representation of a wafer datum.}
\label{fig:spec_data}
\end{figure}
Figure~\ref{fig:spec_data} shows the spectroscopic data for a typical plasma etching process in 3D. There are 2048 wavelengths plotted on the x-axis but it is clear that interesting peaks show up only on a small subset of these. Time is plotted on the y-axis and the height (z-axis) shows the intensity. 

In the analysis presented here the machine learning task is to predict the etch rate from this time-series data (a regression task). In earlier work, this has been done by representing the time series by four summary statistics and using multivariate regression on these statistics \cite{lynn2012global,lynn2012real}. The four statistics are produced by dividing the time series into two stages and representing each stage using mean and standard deviation. The two stages are clearly visible in Figure \ref{fig:spec_data}.


\subsection{The Data} \label{sec:data}
Altogether we have data on the production of \num{1771} wafers. For each of these wafers, we have an estimate of the average etch rate which we take as the dependent variable. As stated already, our ML objective is to predict the etch rate, a regression task. For each wafer, we have the full OES time-series as shown in Figure \ref{fig:spec_data}. This is sampled across \num{2048} wavelengths and contains about 57 ticks for each wavelength.

Rather than consider all \num{2048} wavelengths, in consultation with a domain expert, we reduced this to 38 wavelengths with peaks known to correspond to the key compounds involved in the plasma etching process. 

It is clear from looking at the spectra in Figures \ref{fig:spec_data} that the process has two distinct stages. For this reason, the time-series data can be significantly reduced by simply representing each spectrum time series by four aggregate statistics, the mean and standard deviation in stages 1 and 2. In this way, each wafer can be represented by $2048 \times 4$ data points rather than $2048 \times 60$ data points in the full time-series representation.

\section{Prediction Models}\label{sec:models}
Given our objective of gaining insights from the data we are interested in $k$-Nearest Neighbour regression using DTW on the spectra time-series as the similarity measure. 
The analysis of time-series classification algorithms presented by Bagnall \emph{et al.} \cite{bagnall2017great} shows that 1-NN DTW is an effective method for time-series classification, if not as accurate as state-of-the-art ensemble methods such as COTE. 

We compare $k$-NN DTW against two base-line predictors, multivariate regression on the aggregate statistics and $k$-NN on the aggregate statistics. So we consider three prediction models:
\begin{enumerate}
\item \textbf{R-4M} Multivariate regression using the four aggregate statistics. 
\item \textbf{$k$-NN-4M} $k$-NN regression ($k=5$) using the four aggregate statistics.
\item \textbf{$k$-NN DTW} $k$-NN using DTW ($k=5$) as the similarity measure (see section \ref{sec:DTW} for details).
\end{enumerate}
When these are compared they are always compared on the same underlying wavelengths; for instance in Figure \ref{fig:comparison} they are compared over 1 to 11 wavelengths. 
\subsection{Dynamic Time Warping}\label{sec:DTW}
To find the distance between two data series, the Euclidean formula is a popular choice. But when dealing with time-series data where the series may be displaced in time, the Euclidean distance may be large when the two series are similar, just off slightly on the timeline. To tackle this situation Dynamic Time Warping offers us the malleability of mapping the two data series in a non-linear fashion by warping the time axis \cite{keogh2001derivative}. It creates a cost matrix where the cells contain the distance value of the corresponding data-points and then finds the shortest path through the grid, which minimizes the total distance between them.
\begin{figure}[H]%
    \centering
    \subfloat[Two time series displaced in time.]{{\includegraphics[width=5.65cm]{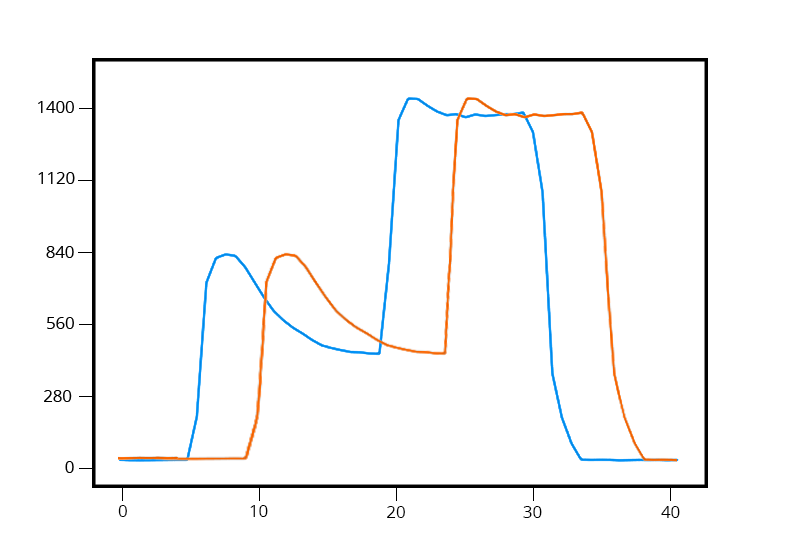} }}%
    \subfloat[Mapping of data-points without warping.]{{\includegraphics[width=5.65cm]{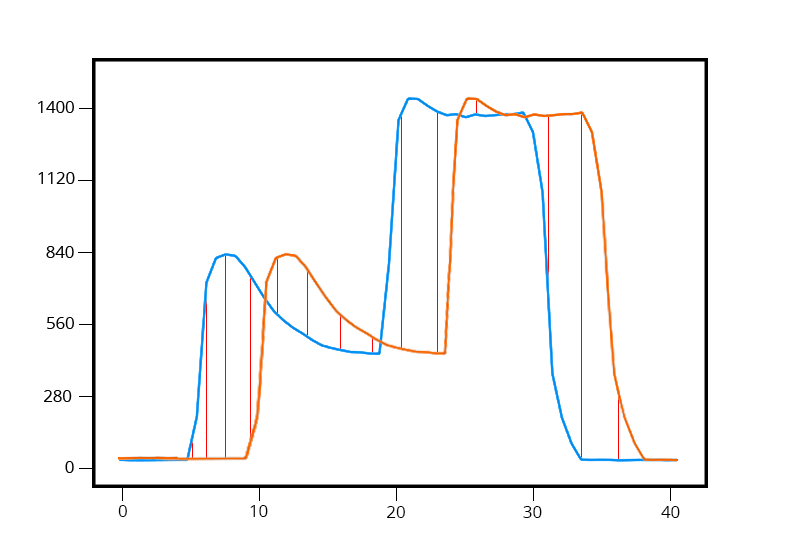} }}%
    \qquad
    \subfloat[Mapping of data-points with warping.]{{\includegraphics[width=5.65cm]{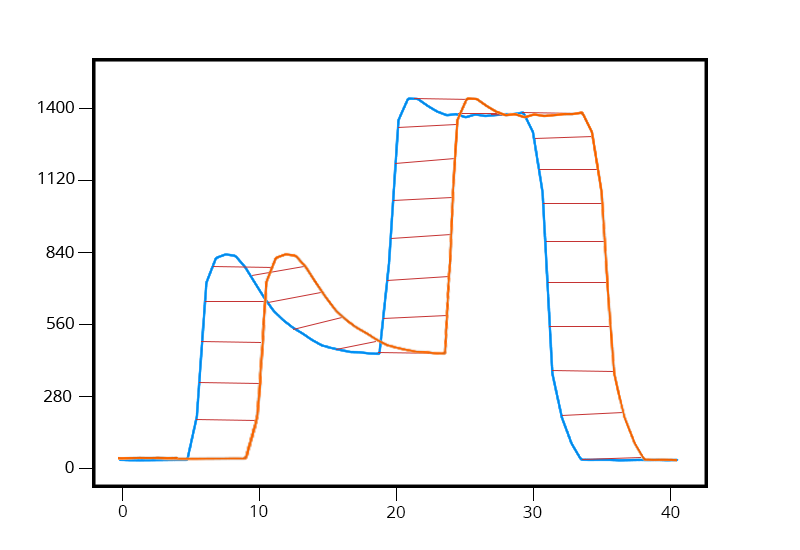} }}%
    \caption{Dynamic Time Warping.}%
    \label{fig:dtw}%
\end{figure}
The figure above illustrates the non-linear mapping done by DTW when we give it a suitable extent of data-points, i.e. max warping window, to consider while fetching the shortest path.
Initially, when we restrict our algorithm with no warping being allowed,  the data points are linearly mapped between the two data series based on the common time axis value. As seen in Figure~\ref{fig:dtw}(b), the algorithm fails to map the trend of a time series data to another. But when we grant the DTW algorithm the flexibility of considering a warping window, the algorithm performs remarkably when mapping the data-points following the trend of the time-series data, which can be visualized in Figure~\ref{fig:dtw}(c). 

\section{Evaluation}\label{sec:eval}
As mentioned in the Introduction, we are more concerned with gaining insights into the factors that influence outcomes that the ability to predict the outcomes themselves. For this reason, we are evaluating the potential of DTW to reveal systematic patterns in the data.

\subsection{Parameter Selection} 

As stated in section \ref{sec:data}, domain knowledge indicates that the \num{2048} wavelengths can be reduced to \num{38}. Through a preliminary feature selection analysis, this was reduced further to 11 wavelengths that were found to be predictive for at least one of the three models. In the next subsection we report on the impact of these 11 wavelengths on prediction accuracy then we conclude the evaluations with some comments on insights. 

\begin{figure}[H]%
    \centering
    \subfloat[$k$-NN-4M performance when trained on 4 summary metrics of a single wavelength against different neighbourhood sizes.]{{\includegraphics[width=5.65cm]{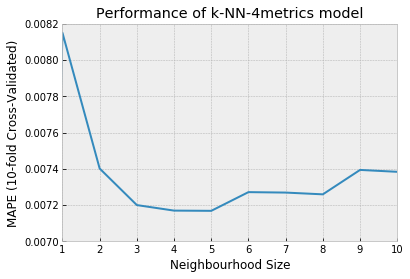} }}%
    \qquad
    \subfloat[$k$-NN-DTW performance when trained on a single wavelength against different warping-window sizes.]{{\includegraphics[width=5.65cm]{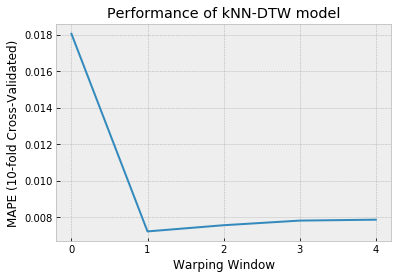} }}%
    \caption{Effect on the performance of $k$-Nearest Neighbour models with different parameters.}%
    \label{fig:parameter}%
\end{figure}

For $k$-NN neighbourhood size ($k$) is a crucial parameter governing performance. Fig. \ref{fig:parameter}(a)  shows the impact of different values of $k$ for the  $k$-NN-4M model. Given that the best performance is achieved when $k=5$ we use this for both $k$-NN models. 

For DTW the size of the warping window has an impact \cite{ratanamahatana_keogh_2005} so we tried different values between 0 and 4. Fig. \ref{fig:parameter}(b) shows the impact of the warping-window on the performance of $k$-NN-DTW model trained on a single wavelength. The model performs the worst when we set the warping-window size to be 0, which is a simple Euclidean distance calculation between the two time-series data. The model performs significantly better when we allow warping but the performance does not improve above a warping window size of one so that parameter value was chosen.

\subsection{Prediction Accuracy} 

\begin{figure}[h]
\centering \includegraphics[width=9cm]{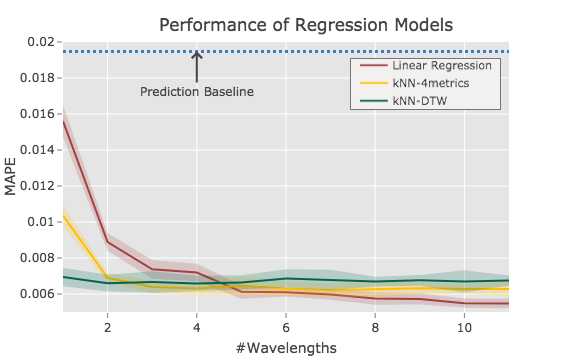}
\caption{Comparison between different regression models. The shading shows the standard deviation.}
\label{fig:comparison}
\end{figure}

Figure \ref{fig:comparison} summarizes the results of a 10-fold cross-validation analysis of the three prediction models using the 11 wavelengths. For each wavelength, the $k$-NN-DTW model has access to the full time-series representation and the other two models use the four summary statistics only. 

Unsurprisingly, when only a single wavelength is considered, the Mean Absolute Percentage Error (MAPE) of the DTW model is significantly better than the models using the four summary statistics. However, as more wavelengths are added the performance of the $k$-NN-DTW model does not improve and the other models close the gap. 
If prediction accuracy were our core objective then it seems that DTW does not offer any benefits because a feature vector based on summary statistics from a wider set of wavelengths provides accuracy that is even slightly better. The results in Figure \ref{fig:comparison} suggest that $k$-NN with four summary statistics from 11 wavelengths (44 features) has the lowest MAPE score. 

This might be interpreted as a variation on the modern wisdom that ``\emph{more data is better than smarter algorithms}" \cite{halevy2009unreasonable}. It is a variation in the sense that the benefits of big data are understood to derive from a large number of examples. Whereas in our case the advantage comes from more extensive data on each example. If summary statistics are available across many wavelengths there is no benefit in using DTW (the smart algorithm in this case).

\subsection{Insights}
The objective in using DTW to identify nearest neighbours was to see if an interpretable measure of similarity would offer insights into the factors influencing outcomes. To test this we selected two wafers, one with the highest etch rate and one with the lowest etch rate and examined their nearest and furthest neighbours as measured by DTW on the first wavelength in Figure \ref{fig:comparison} (564.96nm). 

\begin{figure}[H]%
    \centering
    \subfloat[Wafer with highest etch rate.]{{\includegraphics[width=5.65cm]{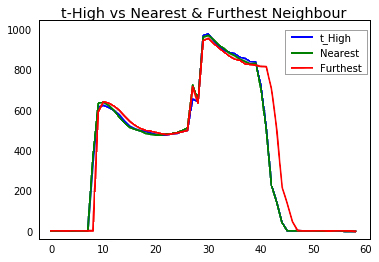} }}%
    \qquad
    \subfloat[Wafer with lowest etch rate.]{{\includegraphics[width=5.65cm]{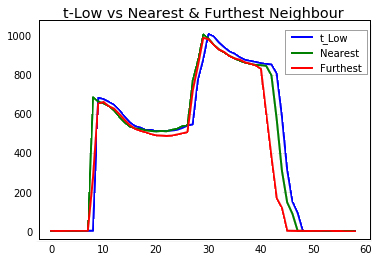} }}%
    \caption{Nearest and furthest neighbours of a wafer based on a single informative wavelength.}%
    \label{fig:insight}%
\end{figure}

Looking at the wafer with the high etch rate first (Figure \ref{fig:insight} (a)) we see that it matches its nearest and furthest neighbours well all along the time series until the end of the second phase where the intensity of the wavelength starts to drop off. It is clear that the nearest neighbour matches perfectly here while this region of the time series accounts for the main difference between the query case and it's most distant neighbour. 
When we turn to the wafer with the low etch rate (Figure \ref{fig:insight} (b)) it appears that the same region of the time-series seems also to account for the difference between the query case and the most different neighbour. This supports the conclusion that this phase of the process is the source of variability in the etch rate. 

\section{Conclusions \& Future Work}\label{sec:conc}

While the work reported here is about predictive analytics the motivation is more to discover insights into factors that influence outcomes rather than generating very accurate predictions. For this reason, we focus on $k$-NN DTW as a model that is interpretable. 

While this research is still ongoing we report two findings here:
 
\begin{itemize}
\item Using DTW does improve prediction accuracy over models that use summary statistics of the time series on a single wavelength. However, when other wavelengths are considered the benefits of using DTW are no longer apparent. 
\item DTW does seem to offer interpretability benefits through the analysis of nearest and furthest neighbours to identify regions of the time series where differences are apparent. 
\end{itemize}

\section{ Acknowledgments}
This publication has resulted from research supported in part by a  grant from Science Foundation Ireland (SFI) under Grant Number 16/RC/3872 and is co-funded under the European Regional Development Fund.

\bibliography{aaltd-DTW} \bibliographystyle{splncs03}

\begin{thebibliography}{1}
\providecommand{\url}[1]{\texttt{#1}}
\providecommand{\urlprefix}{URL }

\bibitem{bagnall2017great}
Bagnall, A., Lines, J., Bostrom, A., Large, J., Keogh, E.: The great time
  series classification bake off: a review and experimental evaluation of
  recent algorithmic advances. Data Mining and Knowledge Discovery  31(3),
  606--660 (2017)

\bibitem{flamm1981design}
Flamm, D.L., Donnelly, V.M.: The design of plasma etchants. Plasma Chemistry
  and Plasma Processing  1(4),  317--363 (1981)

\bibitem{halevy2009unreasonable}
Halevy, A., Norvig, P., Pereira, F.: The unreasonable effectiveness of data.
  IEEE Intelligent Systems  24(2),  8--12 (2009)

\bibitem{keogh2001derivative}
Keogh, E.J., Pazzani, M.J.: Derivative dynamic time warping. In: Proceedings of
  the 2001 SIAM International Conference on Data Mining. pp. 1--11. SIAM (2001)

\bibitem{lynn2012real}
Lynn, S.A., MacGearailt, N., Ringwood, J.V.: Real-time virtual metrology and
  control for plasma etch. Journal of Process Control  22(4),  666--676 (2012)

\bibitem{lynn2012global}
Lynn, S.A., Ringwood, J., MacGearailt, N.: Global and local virtual metrology
  models for a plasma etch process. IEEE Transactions on Semiconductor
  Manufacturing  25(1),  94--103 (2012)

\bibitem{macgearailt2015process}
MacGearailt, N.: Process diagnostics of industrial plasma systems. Ph.D.
  thesis, Dublin City University (2015)

\bibitem{ratanamahatana_keogh_2005}
Ratanamahatana, C.A., Keogh, E.: Three myths about dynamic time warping data
  mining. Proceedings of the 2005 SIAM International Conference on Data Mining
  p. 506–510 (2005)

\bibitem{shul1997high}
Shul, R., McClellan, G., Briggs, R., Rieger, D., Pearton, S., Abernathy, C.,
  Lee, J., Constantine, C., Barratt, C.: High-density plasma etching of
  compound semiconductors. Journal of Vacuum Science \& Technology A: Vacuum,
  Surfaces, and Films  15(3),  633--637 (1997)

\end{thebibliography}

\end{document}